\documentclass{article}
\usepackage{kr}

\usepackage{times}
\usepackage{soul}
\usepackage{url}
\usepackage[hidelinks]{hyperref}
\usepackage[utf8]{inputenc}
\usepackage[small]{caption}
\usepackage{graphicx}
\usepackage{amsmath}
\usepackage{amsthm}
\usepackage{booktabs}
\usepackage{algorithm}
\usepackage{algorithmic}
\usepackage{latexsym}

\usepackage{todonotes}
\usepackage{xargs}
\usepackage{xcolor}
\usepackage{multirow}
\usepackage{amssymb}

\newcommand{\EC}[1]{\textcolor{red}{#1}}

\newboolean{arxivversion}
\setboolean{arxivversion}{true}

\newcommand{\Args}{\mathcal{X}}
\newcommand{\ArgsP}{\Args_p}
\newcommand{\ArgsS}{\Args_s}
\newcommand{\Atts}{\mathcal{A}}

\newcommand{\Supps}{\mathcal{S}}
\newcommand{\BS}{\ensuremath{\tau}}
\newcommand{\SF}{\ensuremath{\sigma}}
\newcommand{\Debate}{\mathcal{D}}
\newcommand{\Proposals}{\mathcal{P}}

\newcommand{\QBAF}{\mathcal{Q}}
\newcommand{\Pros}{\ensuremath{\mathsf{pro}}}
\newcommand{\Cons}{\ensuremath{\mathsf{con}}}
\newcommand{\argpaths}{\mathsf{paths}}

\newtheorem{definition}{Definition}

\newtheorem{metric}{Metric}

\title{Evaluating LLM-Driven Summarisation of Parliamentary Debates \\
with Computational Argumentation}

\author{%
Eoghan Cunningham$^1$\and
James Cross$^2$ \and 
Derek Greene$^1$ \And
Antonio Rago$^3$
\affiliations
$^1$School of Computer Science, University College Dublin\\
$^2$School of Politics and International Relations, University College Dublin\\
$^3$Department of Informatics, King's College London\\
\emails
$^{1,2}$\{firstname, surname\}@ucd.ie,
$^{3}$antonio.rago@kcl.ac.uk
}

\begin{document}

\maketitle

\begin{abstract}

Understanding how policy is debated and justified in parliament is a fundamental aspect of the democratic process. However, the volume and complexity of such debates mean that outside audiences struggle to engage. Meanwhile, Large Language Models (LLMs) have been shown to enable automated summarisation at scale. While summaries of debates can make parliamentary procedures more accessible, 
evaluating whether these summaries faithfully communicate argumentative content remains challenging. Existing automated summarisation metrics
have been shown to correlate poorly with human judgements of \textit{consistency} (i.e., faithfulness or alignment between summary and source). In this work, we propose a formal framework for evaluating parliamentary debate summaries that grounds argument structures in the contested proposals up for debate. Our novel approach, driven by computational argumentation, focuses the evaluation on 
argumentative metrics concerning the faithful preservation of the reasoning presented to justify or oppose policy outcomes. We demonstrate our methods using debates from the European Parliament and associated LLM-driven summaries. 

\end{abstract}


\section{Introduction}
\label{sec:introduction}

Parliamentary debates generate extensive content that must be made comprehensible to citizens, researchers, and policymakers. 
Debate summaries can make the policy-making process more accessible to those who might not engage with parliamentary activity otherwise \cite{tsfati2003debating}, and more understandable for those who do engage \cite{hwang2007applying}. 
Automated summarisation methods are routinely applied to this task \cite{cunningham2025identifying}, and the introduction of modern Large Language Models (LLMs) has meant that convincing summaries can be generated at scale. However, key challenges remain in evaluating these summarisation methods, particularly assessing faithfulness -- sometimes referred to as ``factual consistency'' \cite{fabbri2021summeval}. Human evaluations of LLM-driven summaries of argumentative texts have found that they ``\textit{lack in dimensions that require logical understanding}'' \cite{li2024which}. This is problematic when LLMs are placed between citizens and their representatives. 
Since such evaluations are not scalable, many studies rely instead on automated measures of lexical overlap or semantic similarity \cite{li2024which,roush2024opendebateevidence}. These methods often correlate poorly with human judgments of factual consistency \cite{kryscinski2019neural}. 
In this paper, we show how formal argumentation frameworks \cite{handbook} can 
can help to remedy this issue by providing a structured, interpretable basis for assessing the faithfulness of debate summaries, following existing work in this area \cite{Hautli-Janisz_22}.

We propose an argumentation framework (see §\ref{sec:formalisation}) to represent the arguments made by politicians to oppose or justify specific policy outcomes; i.e., the key provisions contested in the debate. This serves two purposes. 
Firstly, it focuses summary evaluations on outcomes, because promoting parliamentary transparency and democratic participation necessitates summaries that faithfully communicate the reasoning presented to justify or undermine these outcomes.
Secondly, these root proposals provide the shared anchor points for aligning argumentation frameworks from source debates with those extracted from their summaries. Following this alignment, we evaluate the faithfulness of a debate summary on the basis of five 
argumentative metrics introduced in §\ref{sec:formalisation}.

We ground this evaluation setting with a case-study of debates (and associated LLM-driven summaries) from the European Parliament (EP), which provides a proof-of-concept of our approach. Firstly, to demonstrate how these frameworks can be constructed at scale, in §\ref{ssec:argument_mining} we benchmark state-of-the-art argument mining methods on an annotated dataset of arguments from EP debates. Secondly, in §\ref{ssec:evaluation}, we apply our methods (alongside established metrics from Natural Language Processing (NLP)
to a sample of debates to show how our proposed 
metrics can provide a formal and interpretable basis for evaluating debate summaries.



\section{Preliminaries}
\label{sec:preliminaries}

A \emph{quantitative bipolar argumentation framework} (QBAF) \cite{Amgoud_18,Baroni_18} is a quadruple $\langle \Args, \Atts, \Supps, \BS \rangle$ where: $\Args$ is a finite set of \emph{arguments}; $\Atts \subseteq \Args \times \Args$ is a binary relation of (\emph{direct}) \emph{attack}; $\Supps \subseteq \Args \times \Args$ is a binary relation of (\emph{direct}) \emph{support}; $\Atts$ and $\Supps$ are disjoint; and $\BS: \Args \rightarrow [0,1]$ gives the \emph{base score} of an argument.
For the remainder of this section we assume as given a generic QBAF $\QBAF = \langle \Args, \Atts, \Supps, \BS \rangle$.
For any argument $x_i \in \mathcal{X}$, we denote $\Atts(x_i) = \{x_j \in \Args | (x_j, x_i) \in \Atts \}$ as $x_i$'s \emph{attackers} and $\Supps(x_i) = \{x_j \in \Args | (x_j, x_i) \in \Supps \}$ as $x_i$'s \emph{supporters}.
Arguments may be assigned \emph{strengths} representing their acceptability by means of a \emph{gradual semantics} $\SF$, such that for 
$x_i \in \Args$, $\SF(\QBAF,x_i) \in [0,1]$ is calculated by starting from its initial strength (i.e., its base score), which is then recursively increased towards 1 or decreased towards 0 based on the strengths of its attackers or supporters, respectively.
Such gradual semantics satisfy properties that are 
intuitive for representing debates \cite{Tarle_22,Rago_23}.
We use the \emph{DF-QuAD semantics} \cite{Rago_16}, whereby 
for any $x_i \in \Args$, 
    $\SF(\QBAF,x_i) = c(\BS(x_i),\Sigma(\SF(\QBAF, \Atts(x_i))),\Sigma(\SF(\QBAF,\Supps(x_i))))$ 
where, for any $S \subseteq \Args$, $\SF(\QBAF,S)\!\!=\!\!(\SF(\QBAF,x_1),\ldots,\SF(\QBAF,x_m))$ for $(x_1,\ldots,x_m)$, an arbitrary permutation of $S$. Then, 
    %
    $\Sigma$ is such that $\Sigma(())=0$, where $()$ is an empty sequence, and, for $v_1,\ldots,v_n \in [0,1]$ ($n \geq 1$): 
    if $n=1$, then $\Sigma((v_1))=v_1$; if $n=2$, then $\Sigma((v_1,v_2))= v_1 + v_2 - v_1\cdot v_2$; and 
    if $n>2$, then $\Sigma((v_1,\ldots,v_n)) = \Sigma (\Sigma((v_1,\ldots, v_{n-1})),v_n)$. Finally, 
    %
    $c$ is such that, for $v^0,v^-,v^+ \in [0,1]$:
    if $v^-\geq v^+$, then $c(v^0,v^-,v^+)=v^0-v^0\cdot| v^+ - v^-|$; and
    if $v^-< v^+$, then $c(v^0,v^-,v^+)=v^0+(1-v^0)\cdot| v^+ - v^-|$.
    %
For any $x_i, x_j \in \Args$, we let a \emph{path} from $x_i$ to $x_j$ be defined as a sequence $(x_0,x_1), \ldots, (x_{n-1}, x_{n})$ for some $n>0$
, where $x_0 = x_i$, $x_n = x_j$ and, for any $1 \leq k \leq n$, $(x_{k-1}, x_{k}) \in \Atts \cup \Supps$.
Then, $\argpaths(\QBAF,x_i,x_j)$, denotes the set of all paths between any $x_i, x_j \in \Args$. With a small abuse of notation, we treat paths as sets of pairs. The \emph{pro} and \emph{con arguments} \cite{Rago_23} of any $x_i \in \Args$ are $\Pros(\QBAF,x_i) = \{ x_j \in \Args | \exists p \in \argpaths(\QBAF,x_j,x_i) : | p \cap \Atts | \text{ is even} \}$ and $\Cons(\QBAF,x_i) = \{ x_j \in \Args | \exists p \in \argpaths(\QBAF,x_j,x_i): | p \cap \Atts | \text{ is odd} \}$, respectively.

\emph{Bipolar argumentation frameworks} (BAFs)~\cite{Cayrol_05} are triples $\langle \Args, \Atts, \Supps \rangle$, i.e., QBAFs without base scores.
For any $x_i, x_j \in \Args$ such that there is a path from $x_i$ to $x_j$, $(x_1, x_2), \ldots, (x_{n-1}, x_n)$, where $n \geq 3$, this path is: an \emph{indirect attack} from $x_i$ to $x_j$ iff $(x_1, x_2) \in \Atts$ and $(x_k, x_{k+1}) \in \Supps$ $\forall k \in \{ 2, \ldots, n - 1 \}$;
a \emph{supported attack} from $x_i$ on $x_j$ iff $(x_{n-1}, x_n) \in \Atts$ and $(x_k,x_{k+1}) \in \Supps$ $\forall k \in \{ 1, \ldots, n - 2 \}$; or
an \emph{indirect support} from $x_i$ on $x_j$ iff $(x_k,x_{k+1}) \in \Supps$ $\forall k \in \{ 1, \ldots, n - 1 \}$.
A set of arguments $X \subseteq \Args$, also called an \emph{extension}, is said to \emph{set-attack} any $x_i \in \Args$ iff there exists an attack (whether direct, indirect or supported) from some $x_j \in X$ on $x_i$. 
Meanwhile, $X$ is said to \emph{set-support} any $x_i \in \Args$ iff there exists a (direct or indirect) support from some $x_j \in X$ on $x_i$.
Then, a set $X \subseteq \Args$ \emph{defends} any $x_i \in \Args$ iff $\forall x_j \in \Args$, if $\{ x_j \}$ set-attacks $x_i$ then $\exists x_k \in X$ such that $\{ x_k \}$ set-attacks $x_j$.
Any set $X \subseteq \Args$ is then said to be \emph{conflict-free} iff $\nexists x_i, x_j \in X$ such that $\{ x_i \}$ set-attacks $x_j$.
The notion of a set $X \subseteq \Args$ being \emph{d-admissible} (based on \emph{admissibility} in \cite{Dung_95}) requires that $X$ is conflict-free and defends all of its elements. 
Finally, $X$ is \emph{d-preferred} iff it is d-admissible and maximal wrt set-inclusion. We let $\Sigma_d(\QBAF)$ denote the set of all d-preferred extensions in $\QBAF$ and we use \emph{credulous}, rather than \emph{sceptical}, acceptance of arguments, i.e., to be accepted they must be in one extension, rather than all extensions. 


\section{Formally Representing Parliamentary Debates and Summaries}
\label{sec:formalisation}

In this section, we first introduce a formal representation of parliamentary debates, which may be derived from the debate transcipts themselves, or summaries thereof. Then, we introduce 
argumentative metrics for individual QBAFs and for pairwise comparisons between two QBAFs from the two sources, thus providing a means of assessing summarisation methods formally.
We represent the debates as follows.

\begin{definition}
\label{def:ourBAF}
	Given a debate $\Debate$ on proposal $\Proposals$ containing $n$ provisions, $\Proposals = \{p_1, \ldots, p_n \}$, a \emph{QBAF representing $\Debate$} is a QBAF $\langle \Args, \Atts, \Supps, \BS \rangle$ such that:
	\begin{itemize}
		\item $\Args = \ArgsP \cup \ArgsS$, where $\ArgsP \cap \ArgsS = \emptyset$ and:
        \begin{itemize}
            \item $\ArgsP$ is the set of \emph{proposal arguments}  where $ |\ArgsP| = |\Proposals|$ and $\forall p_i \in \Proposals$, where $i \in \{ 1, \ldots, n \}$, $\exists x_i \in \ArgsP$;
            \item $\ArgsS$ is the set of \emph{speech arguments};
        \end{itemize}
        \item $\Atts \subseteq \ArgsS \times (\ArgsS \cup \ArgsP)$ and $\Supps \subseteq \ArgsS \times (\ArgsS \cup \ArgsP)$;
        \item $\BS$ is such that $\forall x_i \in \ArgsP$, $\BS(x_i) = 0.5$.
	\end{itemize}
\end{definition}

\begin{figure}[t]
    \centering
    \includegraphics[width=\linewidth]{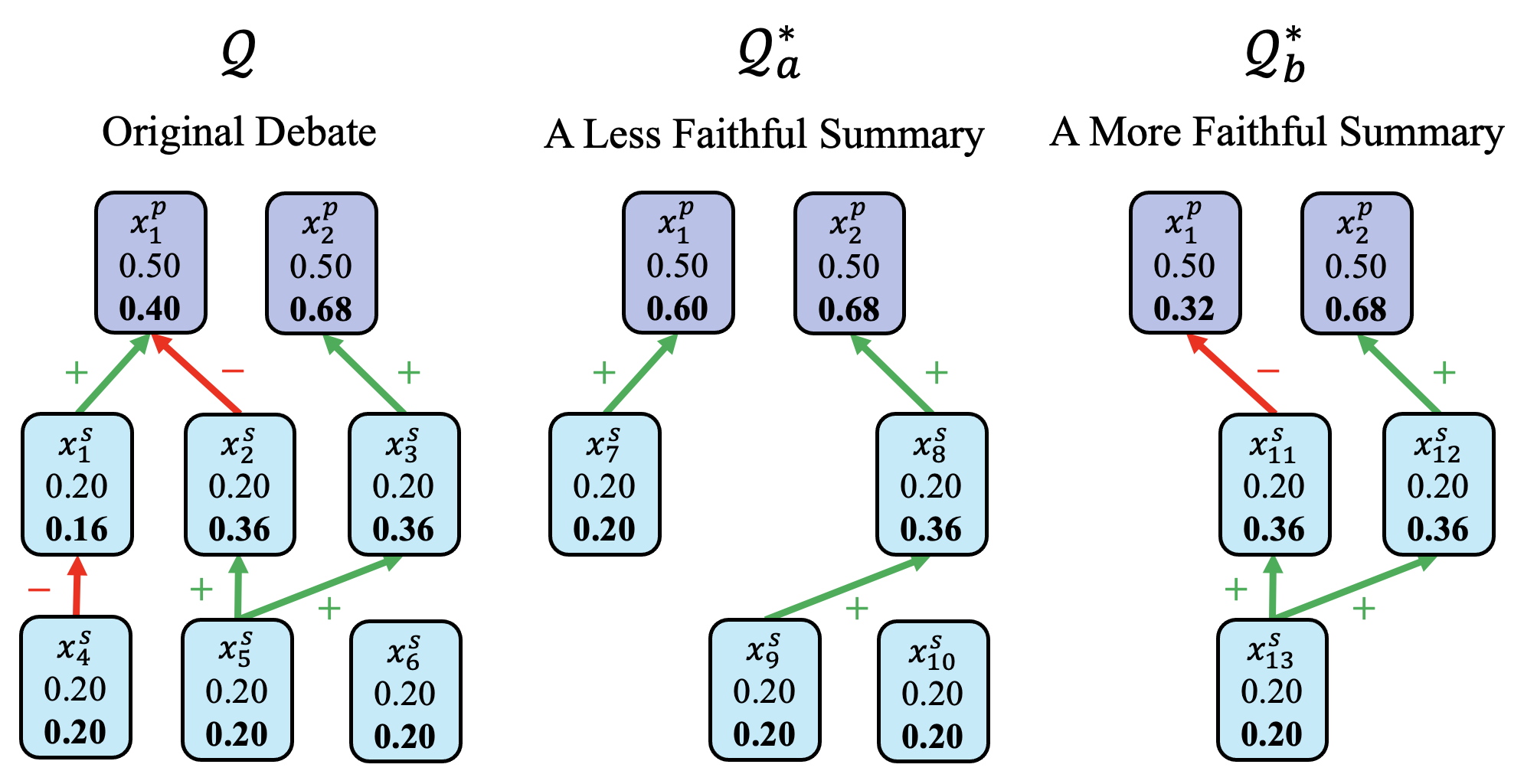}
    \caption{Three example QBAFs representing a debate on two proposals $\Proposals = \{ p_1, p_2 \}$, where $\QBAF$ may be taken to be extracted from the original debate text, and $\QBAF_a^*$ and $\QBAF_b^*$ from summaries thereof. Proposal (speech) arguments are represented by purple (blue, respectively) nodes with superscript $p$ ($s$, respectively), attacks by red edges labelled ``$-$'' and supports by green edges labelled ``$+$''. Arguments' base scores (strengths, calculated by the DF-QuAD semantics) are the values in normal (bold, respectively) text. 
    }
    \label{fig:example_figure}
\end{figure}

Our QBAFs thus represent all $n$ provisions being debated as proposal arguments, while speech arguments are those put forward by speakers taking part in the debate. Intuitively, attacks and supports come from speech arguments only, and they are directed towards proposal arguments or other speech arguments. Regarding base scores, we fix those of proposal arguments to be the midpoint to allow for the balance between attacking and supporting speech arguments to determine the proposal arguments' strengths. Meanwhile, base scores of speech arguments are left undetermined so that they can be tailored to specific applications, e.g. some may assume a constant base score, while in others, modelling argument similarity may allow the base score to vary using argument prominence, i.e., the number of times the argument was put forward. In our later examples, we assume a base score of $0.2$ for all arguments in $\ArgsS$, as this gave reasonable results in our preliminary experimentation, and we calculate strength scores using the \textit{DF-QuAD} semantics.\footnote{In the supplementary material (SM)\ifthenelse{\boolean{arxivversion}}{}{ of the technical report 
for this paper \cite{Cunningham_26X}}, we report comparable results with other base scores revealing minimal effect on interpretations.}
We leave experimentation with other (constant or varying) base scores to future work. Examples of our QBAFs are shown in Figure \ref{fig:example_figure} and are assessed along our 
metrics, which we introduce next, in Table \ref{tab:example}.


For our first 
metric, we consider the relevance of the speech arguments by measuring how many 
are connected to at least one
proposal argument. Speech arguments without a path to the proposal arguments are thus deemed as irrelevant.


\begin{metric}
\label{prop:relevance}
    The \emph{argument relevance} of any $\QBAF = \langle \Args, \Atts, \Supps, \BS \rangle$ representing a debate $\Debate$ which concerned proposal $\Proposals = \{p_1, \ldots, p_n \}$ is 
    $\pi_{ar}(\QBAF) = \frac{|\{ x_i \in \ArgsS | \exists x_j \in \ArgsP : \argpaths(x_j,x_i) \neq \emptyset \}|}{|\ArgsS|}.$
\end{metric}

As shown in Table \ref{tab:example}, $\QBAF$ and $\QBAF_a^*$ do not have maximal relevance due to the unconnected argument in each, whereas this is not the case for $\QBAF_b^*$. High relevance may be suitable for some summaries, but in others, maintaining the same relevance as the original debate may be more important, while such ``irrelevant'' arguments have also been shown to be helpful in argumentative analysis \cite{Ruiz-Dolz_25}.

Next, we consider, within the set of relevant speech arguments, the balance of evidence for and against a given proposal argument. Here, we measure the number of relevant speech arguments \textit{for} a proposal argument, normalised by the sum of those \textit{for} and \textit{against} it. 

\begin{metric}
\label{prop:procon}
    The \emph{pro-con ratio} of any $x_i \in \ArgsP$, where $\QBAF = \langle \Args, \Atts, \Supps, \BS \rangle$ represents a debate $\Debate$ which concerned proposals $\Proposals = \{p_1, \ldots, p_n \}$, is 
    $\pi_{pcr}(\QBAF, x_i) = \frac{ | 
    \Pros(\QBAF,x_i) 
    | }{ | 
    \Pros(\QBAF,x_i) \cup \Cons(\QBAF,x_i) 
    | }.$    %
    %
\end{metric}

In Table \ref{tab:example}, $\QBAF_b^*$ maintains the same pro-con ratios as $\QBAF$, which we believe to be desirable in a summary, while $\QBAF_a^*$ does not due to con arguments of $x_1^p$ not being included. 

We now turn to the pairwise 
metrics. For the remainder of this paper, we assume as given two QBAFs $\QBAF = \langle \Args, \Atts, \Supps, \BS \rangle$ and $\QBAF^* = \langle \Args^*, \Atts^*, \Supps^*, \BS^* \rangle$ representing a debate $\Debate$ concerning 
proposal $\Proposals = \{p_1, \ldots, p_n \}$. We assume that $\QBAF$ and $\QBAF^*$ were extracted from the original debate text and a summarisation thereof, respectively.



We consider pairwise 
metrics 
for proposal arguments only. While our framework supports speech argument 
metrics (see the SM\ifthenelse{\boolean{arxivversion}}{}{ in \cite{Cunningham_26X}}), these will require the detection of similar 
arguments across the two frameworks, which we leave to future work.





Our first pairwise 
metric checks how many of the proposal arguments' acceptance statuses (by the d-preferred semantics) remained consistent; giving an indication of whether extension-based semantics' results are preserved.


\begin{metric}
\label{prop:preferability}
    The \emph{d-preferability-consistency} of $\QBAF^*$ wrt $\QBAF$ is $\pi_{dpc}(\QBAF^*,\QBAF) = \frac{|S|}{|\ArgsP^*|}$, where $S \subseteq 
    \ArgsP^*$ is the maximal subset such that $\forall x_i \in S$, $\exists X \in \Sigma_d(\QBAF)$ where $x_i \in X$ iff $\exists X^* \in \Sigma_d(\QBAF^*)$ where $x_i \in X^*$.
\end{metric}


Table \ref{tab:example} shows that $\QBAF_b^*$ preserves the acceptability statuses of both arguments, while $\QBAF_a^*$ does not for $x_1^p$ since both of its attackers were excluded from this summary.

\begin{table}[t]
    \begin{center}
    \begin{tabular}{ccccc}
    \toprule
    &
    $\QBAF$ &
    $\QBAF_a^*$  & 
    $\QBAF_b^*$ \\[1pt]
    \hline
    %
    $|\ArgsS|$ &
    6 & 
    4 &
    3 \\
    %
    $\pi_{ar}(\cdot)$ & 
    0.83 & 
    0.75 & 
    1.00 \\
    %
    $\pi_{pcr}(\cdot,r_1^p)$ & 
    0.25 & 
    1.00 & 
    0.00 \\
    %
    $\pi_{pcr}(\cdot,r_2^p)$ & 
    1.00 & 
    1.00 & 
    1.00 \\
    %
    $\pi_{dpc}(\cdot,\QBAF)$ & 
    --- & 
    0.50 & 
    1.00 \\ 
    %
    $\pi_{bc}(\cdot,\QBAF)$ & 
    --- & 
    0.50 & 
    1.00 \\ 
    %
    $\pi_{\epsilon a}(\cdot,\QBAF)$ &
    --- & 
    0.50 & 
    1.00 \\ 
    \bottomrule
    \end{tabular}
    \end{center}
    \protect\caption{The QBAFs from Figure \ref{fig:example_figure} assessed along our argumentative metrics. Note 
    that $\epsilon = 0.1$ was used for 
    $\pi_{\epsilon a}$.} 
    \label{tab:example}
    \end{table}


Our final two 
metrics concern gradual semantics, so we assume a given $\SF$.
Metric~\ref{prop:balanced} checks whether proposal arguments that received net support (have strength greater than $
0.5$) or net opposition (have strength less than $
0.5$) remain so in the debate summary, and is inspired by the \emph{balance} property in \cite{Baroni_18}. This important 
metric ensures that any summary of a debate remains faithful to, and does not misrepresent, the balance of the discussion, whether in favour of or against the each of the proposals.



\begin{metric}
\label{prop:balanced}
    The \emph{balance-consistency} of $\QBAF^*$ wrt $\QBAF$ is $\pi_{bc}(\QBAF^*,\QBAF) = \frac{|S|}{|\ArgsP^*|}$, where $S \subseteq 
    \ArgsP^*$ is the maximal subset such that $\forall x_i \in S$, $\SF(\QBAF^*, x_i) > 0.5$ iff $\SF(\QBAF, x_i) > 0.5$; and $\SF(\QBAF^*, x_i) < 0.5$ iff $\SF(\QBAF, x_i) < 0.5$.
\end{metric}


Table \ref{tab:example} shows that the strengths in $\QBAF_b^*$ are consistent with those in $\QBAF$, while, again, $\QBAF_a^*$
performs poorly in the metric for $x_1^p$ since its strength reflects net support, rather than the net opposition it receives in $\QBAF$.

Our final 
metric is a little stricter, checking the number of proposal arguments in the summary that maintain their original strength within a specified error, assessing the accuracy of the summarised proposal arguments' strengths.
While we expect some variance in these strengths given that a summary cannot contain all arguments, the error $\epsilon$ should be selected carefully to provide suitable limits.


\begin{metric}
\label{prop:epsilon}
    The \emph{$\epsilon$-accuracy} of $\QBAF^*$ wrt $\QBAF$ is $\pi_{\epsilon a}(\QBAF^*,\QBAF) = \frac{|S|}{|\ArgsP^*|}$, where $S \subseteq 
    \ArgsP^*$ is the maximal subset such that $\forall x_i \in S$, $\SF(\QBAF^*, x_i) \in [\SF(\QBAF, x_i) - \epsilon, \SF(\QBAF, x_i) + \epsilon]$.
\end{metric}

Table \ref{tab:example} again shows that the strengths in $\QBAF_b^*$ are within $\epsilon = 0.1$ of those in $\QBAF$, while $x_1^p$ 
performs worse for $\QBAF_a^*$ since its strength here has changed by more than $\epsilon$. 

Our running example has thus illustrated that while $\QBAF_b^*$ gives a more compact summary of $\QBAF$, by our argumentative 
metrics it provides a much more faithful summary.

\section{Can LLMs Summarise European Parliament Debates?}
\label{sec:evaluation}

In this section, we show how our 
approach can be used to evaluate summarisation of parliamentary debates at scale
.


We use a dataset of 511 arguments across three EP debates from 2006--2008. Each debate contains between 107 and 198 arguments. The debates follow a structured process where Members of the European Parliament (MEPs) discuss individual policy provisions in a proposal before voting. Each resolution 
has numbered paragraphs stating provisions.
For example,
\textit{``The European Parliament recommends that the mandate of the ECDC be extended to non-communicable diseases.''}
For a given debate, the numbered paragraphs from the debated proposal form the set of 
proposal arguments $\ArgsP$.
We let the arguments identified in the debate transcript represent the 
speech arguments $\ArgsS$. These arguments, and the attack and support relations between them, are mined from EP debates using the approach detailed next. 

\subsection{Argument Mining}
\label{ssec:argument_mining}



While our framework is agnostic of the method of argument mining, it is important to evaluate the extent to which the framework can be mined reliably and automatically from debates in parliament. For this evaluation, we rely on 
\emph{relation-based argument mining} \cite{carstens2015towards}. Here, rather than classifying each sentence in a text as argumentative (or non-argumentative) in isolation, we classify all pairs of sentences according to their relation: \textit{support}, \textit{attack}, or \textit{neither}. Thus, we define a sentence as relevant (or \textit{argumentative}) if it attacks or supports an existing argument.  

We include a temporal constraint in the mining of $\QBAF$, such that $(\Atts \cup \Supps) \cap (\ArgsS \times \ArgsS) \subseteq \{(x_j,x_i) |  j>i\}$, where the subscripts of arguments indicate their temporal order in the debate. That is, speech arguments can only attack or support speech arguments that precede them in the debate. Accordingly, $\QBAF$ contains $n$ multitrees that are built around the root proposals. No such temporal constraint could be assumed for the text of a debate summary; thus, we permit cycles in $\QBAF^*$, and rely on approximate convergence under the DF-QuAD semantics ($\delta = 10^{-3}$).
QBAFs are constructed using Argument Relation Classification (ARC).


To evaluate SOTA models in ARC in this context, we compile a dataset of 481 argument pairs, consisting of 377 arguments. We classify relations into \textit{attack}, \textit{support}, and \textit{neither}, by aggregating annotations across three human coders with expertise in EU politics. Each pair is classified by two coders, and any disagreements are resolved by a third. Cohen's kappa agreement scores between annotators ranged from `fair' ($\kappa$ = 0.30) to `moderate' ($\kappa$ = 0.54), indicating that classifying relations is more challenging for some topics than others. While fair-to-moderate agreement scores are typical in argument mining tasks \cite{lawrence2019argument}, an overall agreement of 76.6\% indicates that the task involves subjective judgement on the part of the annotators, which may reasonably limit the performance ceiling for automated methods.
Following recent studies demonstrating the capabilities of LLMs in \cite{savigny2025amelia,gorur2025can}
, we evaluate four different models on this task in various settings (i.e., zero-shot, few-shot, and fine-tuned). We report their performance in Table~\ref{tab:benchmarking}, with results for additional models evaluated (and all prompts used) in the SM\ifthenelse{\boolean{arxivversion}}{}{ in \cite{Cunningham_26X}}. 

\begin{table}[h]
    \centering
    \setlength{\tabcolsep}{5pt}
    \begin{tabular}{lrr}
   \toprule
Model & Accuracy & F1 \\
\midrule
Claude Sonnet 4.5 (zero-shot) & \textbf{0.74} & \textbf{0.78} \\
Claude Haiku 4.5 (zero-shot) & 0.68 & 0.73 \\
Qwen3 30B (zero-shot) & 0.64 & 0.72 \\
Llama 3.1 8B (fine-tuned) & 0.21 & 0.27 \\
\bottomrule
    \end{tabular}
    \caption{Argument Relation Classification from EP debates. Claude Sonnet and Haiku are proprietary pre-trained LLMs from \textit{Anthropic}. Qwen3 is a open-weight, pre-trained LLM from \textit{Alibaba}. Llama 3.1 (an open-weight model from \textit{Meta}) was fine-tuned on existing 8 ARC datasets by~\protect\cite{savigny2025amelia}.}
    \label{tab:benchmarking}
\end{table}

Critically, we find that larger reasoning models implemented in zero-
shot settings outperform smaller fine-tuned models when applied to new data. 
This aligns with recent studies of LLMs in argumentative tasks, finding that smaller models struggle even with fine-tuning \cite{Furman_23}, and especially compared to larger models \cite{Bezou-Vrakatseli_25}.
We hope to extend our dataset in future work to investigate further whether LLMs really learn to mine arguments, or just the datasets themselves \cite{Feger_25}.
In a trade-off between cost and performance, we use Claude Haiku to construct the QBAFs
in the remainder of our case-study.

\subsection{Summary Evaluation}
\label{ssec:evaluation}

Next, we generate summaries for a sample of EP debates and apply our proposed methods of summary evaluation alongside three existing metrics. \textit{ROUGE} \cite{lin2004rouge} is based on explicit n-gram overlap between the source and summary content. Thus, it is liable to over-penalise fair abstraction or paraphrasing in summaries. \textit{BERTScore} \cite{zhang2019bertscore} accounts for abstraction by measuring semantic similarity between tokens in the source and the summary. However, when comparing large volumes of text, these methods are unable to sufficiently penalise inconsistencies in summaries in the presence of strong signals of semantic similarity. \textit{Natural Language Inference} (NLI) models are designed to address this issue by detecting factual inconsistencies in the summary via \textit{entailment classification}. However,  these metrics evaluate summaries holistically against source text, offering aggregate scores and lacking any formal representation of the argumentation in the debate. As such they obscure which aspects of the debate content are preserved or lost. We list all scores for two sample debates in Table~\ref{tab:evaluation} and all prompts used in summarisation are provided in the SM\ifthenelse{\boolean{arxivversion}}{}{ in \cite{Cunningham_26X}}.


\begin{table}[t]
    \centering
    \setlength{\tabcolsep}{3pt}
    \begin{tabular}{cccc|ccc}
    \toprule
     &
    \multicolumn{3}{c}{Debate 1} &
    \multicolumn{3}{c}{Debate 2} \\
    \cmidrule(lr){2-4} \cmidrule(lr){5-7}
     & 
    $\QBAF$ &
    $\QBAF^*_{\text{Sonnet}}$ &
    $\QBAF^*_{\text{Haiku}}$ &
    $\QBAF$ &
    $\QBAF^*_{\text{Sonnet}}$ &
    $\QBAF^*_{\text{Haiku}}$ \\ [2pt]
    \hline
    R-2 & 
    ---  &
    0.24 &
    0.05 & 
    ---  &
    0.12 & 
    0.04 \\
    BERT &
    ---  & 
    0.82 &
    0.83 &
    ---  &  
    0.84 & 
    0.84 \\
    NLI & 
    --- &  
    0.60 &
    0.69 & 
    --- & 
    0.59 & 
    0.56 \\
    \hline
    $\pi_{ar}(\cdot)$ & 
    1.00 &  
    0.97 &
    0.93 & 
    1.00 &
    0.97 & 
    1.00 \\ 
    $\bar{\pi_{pcr}(\cdot,r_i^p)}$ & 
    0.53 &  
    0.52 &
    0.58 & 
    0.50 & 
    0.52 & 
    0.65 \\
    $\pi_{dpc}(\cdot,\QBAF)$ & 
    --- &   
    1.00 & 
    1.00 & 
    --- &  
    1.00 & 
    1.00 \\
    $\pi_{bc}(\cdot,\QBAF)$ & 
    --- &  
     0.26 & 
     0.19 & 
     --- &  
     0.53 & 
     0.47 \\
    $\pi_{\epsilon a}(\cdot,\QBAF)$ &
    --- &
    0.52 &
    0.30 &
    --- &
    0.40 &
    0.00 \\
    \bottomrule
    \end{tabular}
    \caption{Comparing LLM-generated summaries to original debate transcripts across two sampled debates. We include traditional NLP-based summarisation metrics (ROUGE, BERTScore, Sentence-NLI)
    for comparison with our argumentative metrics. For $\pi_{pcr}$, we report the mean over all provisions $r_i^p \in \ArgsP$.}
    \label{tab:evaluation}
\end{table}

Table~\ref{tab:evaluation} shows traditional metrics confirm the summaries are abstractive (low ROUGE), semantically coherent with the source debates (high BERTScore), with many sentences supported by the source (NLI: 0.56--0.69). However, these metrics provide minimal discrimination between summaries that differ substantially in argument structure. In contrast, our 
metrics reveal clear degradation patterns. Firstly, Haiku-generated summaries fail  to preserve the pro-con 
ratio of either debate ($\pi_{pcb}$).  Second, all summaries struggle to accurately communicate the balance of evidence for and against the key proposals ($\pi_{bc}$ of 0.19--0.53) and largely fail to preserve precise proposal strengths ($\pi_{\epsilon a}$ of 0.00--0.52). 
Critically, according to the Haiku generated summary, none of the provision strengths are within $\epsilon = 0.1$ of the scores in the original debate ($\pi_{\epsilon a}(\QBAF^*_{\text{Haiku}},\QBAF) = 0$).
Meanwhile, 
$\pi_{dpc}$ gives maximum values only because no proposal arguments were in d-preferred extensions, showing that gradual semantics may be more effective for this task. Unsurprisingly, the larger model (Sonnet) scores better across almost all
metrics.

As an illustrative example, Table~\ref{tab:interpretation} shows how a subset of the proposal strengths have been misrepresented in the summaries of the second debate.
The summaries exclude much of the argumentation that undermined the root provisions. Key proposal arguments that were contested in the debate ($\sigma(\QBAF, x_i^p) \approx$ 0.5) are presented as overwhelmingly supported in the summaries ($\sigma(\QBAF^*_{\text{Haiku}}, x_i^p) \gg$ 0.5). 

\begin{table}[]
    \centering
    \setlength{\tabcolsep}{3pt}
    \begin{tabular}{llllllll}
    \toprule
     & $x_{10}^p$ & $x_{11}^p$ & $x_{12}^p$ & $x_{13}^p$ & $x_{14}^p$ & $x_{15}^p$ & $x_{16}^p$ \\
    \midrule
    $\QBAF$ & 0.53 & 0.50 & 0.50 & 0.50 & 0.50 & 0.47 & 0.60 \\
    \midrule
    $\QBAF^*_{\text{Sonnet}}$ & 0.54 & 0.52 & 0.78 & 0.78 & 0.82 & 0.54 & 0.90 \\
    $\QBAF^*_{\text{Haiku}}$ & 1.00 & 0.90 & 0.90 & 0.90 & 0.90 & 0.89 & 0.82 \\
    \bottomrule
    \end{tabular}
    \caption{
    Strengths for a sample of proposal arguments in
    QBAFs extracted from the original debate ($\QBAF$) and LLM summaries ($\QBAF^*$).}
    \label{tab:interpretation}
    \vskip-1.5em
\end{table}

\section{Conclusions and Future Work}
\label{sec:conclusions}
We proposed a formal argumentation framework for evaluating parliamentary debate summaries that represents debates as QBAFs constructed around contested policy outcomes. The framework demonstrates how computational argumentation can focus evaluations on faithfully communicating the motivations for policy outcomes rather than on general semantic coherence or similarity. We achieve this through 
argumentative metrics that assess argument acceptability (and thus, the support for proposed outcomes) from the perspectives of both extension-based and gradual semantics, showing that the latter may be more effectively discriminative.

Future work will develop methods for matching similar arguments across sources and summaries to enable speech argument evaluation beyond our current provision-focused 
metrics. 
This would support precision-recall analysis of argumentative content preservation, and facilitate identification of speech arguments that are omitted or hallucinated. 
Further, it would be interesting to assess the usefulness of our 
metrics via user studies, ensuring that summaries judged by users as being faithful align with our 
argumentative metrics over standard metrics, and also in downstream tasks. Another fruitful line of work could be to investigate whether QBAFs extracted via LLMs, e.g. as in \cite{Freedman_25}, are faithful according to our metrics.
Finally, our approach could be supplemented with existing techniques for argument mining in political debates, e.g. leveraging knowledge graph embeddings \cite{Dore_25}.

\section*{AI Declaration}
The authors did not use generative AI tools in any part of the paper-writing process.

\section*{Acknowledgments}
This work was partially supported by Research Ireland grant number SFI/12/RC/2289 P2.

    %
    %
    %
    %
    %
    %
    %
    %




\bibliographystyle{kr}
\bibliography{bib}

\ifthenelse{\boolean{arxivversion}}{}{\iffalse}

\newpage

\clearpage

\section*{Supplementary Material}






\subsection*{Additional Metrics}

Here we propose additional metrics for the speech arguments in the debate and summary. As mentioned, these properties would require methods for matching speech arguments between the original debate $\ArgsS$ and the summary $\ArgsS^*$, which we leave to future work.

First, if deployed a method to match speech arguments across the QBAFs, we would then be able to measure their precision and recall.

\begin{metric}
    The \emph{speech argument precision} of $\QBAF^*$ wrt $\QBAF$ is $\pi_{sap}(\QBAF^*,\QBAF) = \frac{|\ArgsS \cap \ArgsS^*|}{|\ArgsS^*|}$.
\end{metric}

\begin{metric}
    The \emph{speech argument recall} of $\QBAF^*$ wrt $\QBAF$ is $\pi_{sar}(\QBAF^*,\QBAF) = \frac{|\ArgsS \cap \ArgsS^*|}{|\ArgsS|}$.
\end{metric}

High precision seems more important than high recall in this setting given that we do not want ``hallucinated'' arguments, and some speech arguments will likely be discarded since it is a summary.



Finally, we introduce a measure of consistency in the pro and con statuses of arguments, which seems important to ensure relations remain consistent across the whole QBAF.

\begin{metric}
    The \emph{pro-con consistency} of $\QBAF^*$ wrt $\QBAF$ is $\pi_{pcc}(\QBAF^*,\QBAF) = \frac{|S|}{|\ArgsS \cap \ArgsS^*|}$, where $S \subseteq 
    \ArgsS \cap \ArgsS^*$ is the maximal subset such that $\forall x_i \in S$ and $\forall x_j \in \ArgsP \cap \ArgsP^*$:
    \begin{itemize}
        \item $x_i \in \Pros(\QBAF^*,x_j)$ iff $x_i \in \Pros(\QBAF,x_j)$; and
        \item $x_i \in \Cons(\QBAF^*,x_j)$ iff $x_i \in \Cons(\QBAF,x_j)$.
    \end{itemize}
\end{metric}


Many other metrics concerning the speech arguments could also be useful, e.g. ensuring that all of those with a strength over a certain threshold (i.e., were strongly accepted arguments), or those which have over a certain threshold of attackers and supporters (i.e., generated much discussion), are included in the summary.

\subsection*{Full Experimental Results}

Next, we provide the full set of results from all models evaluated in the ARC task.

\begin{table}[h]
    \centering
    \setlength{\tabcolsep}{5pt}
    \begin{tabular}{lrr}
   \toprule
Model & Accuracy & F1 \\
\midrule
Claude Sonnet 4.5 (zero-shot) & \textbf{0.74} & \textbf{0.78} \\
Claude Haiku 4.5 (few-shot) & 0.69 & 0.75 \\
Claude Haiku 4.5 (zero-shot) & 0.68 & 0.73 \\
Claude Haiku 3.5 (zero-shot) & 0.65 & 0.71 \\
Claude Haiku 3.5 (few-shot) & 0.55 & 0.62 \\
Qwen3 30B (zero-shot) & 0.64 & 0.72 \\
Qwen3 30B (few-shot) & 0.45 & 0.54 \\
Llama 3.1 8B (fine-tuned) & 0.21 & 0.27 \\
\bottomrule
    \end{tabular}
    \caption{Argument Relation Classification from EP debates. Claude Sonnet and Haiku are proprietary pre-trained LLMs from \textit{Anthropic}. Qwen3 is a open-weight, pre-trained LLM from \textit{Alibaba}. Llama 3.1 (an open-weight model from \textit{Meta}) was fine-tuned on existing 8 ARC datasets by~\protect\cite{savigny2025amelia}.}
\end{table}

Finally, we provide measures for our metrics (those that are dependent on gradual semantics) for a variety of chosen base scores for $\ArgsS$.  


\EC{\begin{table}[h]
    \centering
    \setlength{\tabcolsep}{2pt}
    \begin{tabular}{llrr|rr}
\toprule
& & \multicolumn{2}{c}{Debate 1} & \multicolumn{2}{c}{Debate 2} \\
\cmidrule(lr){3-4} \cmidrule(lr){5-6}
$\tau$ & Metric & $\QBAF^*_{\text{Sonnet}}$ & $\QBAF^*_{\text{Haiku}}$ & $\QBAF^*_{\text{Sonnet}}$ & $\QBAF^*_{\text{Haiku}}$ \\
\cmidrule{1-6}
\multirow{2}{*}{0.15}
& $\pi_{bc}(\cdot,\QBAF)$& 0.26 & 0.22 & 0.60 & 0.53 \\
& $\pi_{\epsilon a}(\cdot,\QBAF)$ & 0.37 & 0.30 & 0.40 & 0.00 \\
\cmidrule{1-6}
\multirow{2}{*}{0.2} 
& $\pi_{bc}(\cdot,\QBAF)$ & 0.26 & 0.19 & 0.53 & 0.47 \\
& $\pi_{\epsilon a}(\cdot,\QBAF)$ & 0.52 & 0.30 & 0.40 & 0.00 \\
\cmidrule{1-6}
\multirow{2}{*}{0.25} 
&$\pi_{bc}(\cdot,\QBAF)$ & 0.26 & 0.19 & 0.47 & 0.40 \\
&$\pi_{\epsilon a}(\cdot,\QBAF)$ & 0.59 & 0.26 & 0.53 & 0.00 \\
\bottomrule
\end{tabular}
    \caption{Balance-consistency and $\epsilon$-accuracy scores assuming different base scores for speech arguments. The key takeaways from proof-of-concept example in the paper are unaffected by changes in this initial base score $\tau$. Automatically assigning heterogeneous base scores to speech arguments during argument mining is a natural direction for future work.}
    \label{tab:evaluation_appendix}
\end{table}}

\clearpage

\subsection*{Prompts}
Below are the prompts for debate summarisation and argument mining with LLMs.

\noindent\textbf{Summarisation Prompt}
\begin{center}
    
\noindent\begin{minipage}{0.95\linewidth}
\ttfamily\small
```\\
Please provide a comprehensive summary of the following parliamentary debate transcript.\\

{[}Optional: Title: <debate title>{]}\\

Transcript:\\
<formatted transcript with {[}ID{]} Speaker: text for each intervention>\\

Please provide a summary of the debate. Use complete paragraphs and sentences. Do not use tables or lists. Do not use markdown formatting.\\

Summary:\\
'''
\end{minipage}
\end{center}

\noindent\textbf{ARC System Prompt}

\begin{center}
\noindent\begin{minipage}{0.95\linewidth}
\ttfamily\small
```\\You are an expert in argumentation. Your task is to determine the type of relation between [SOURCE] and [TARGET]. The type of relation would be in the [RELATION] set. Utilize the [TOPIC] as context to support your decision

Your answer must be in the following format with only the type of the relation in the answer section:\\
<|ANSWER|><answer><|ANSWER|>.\\'''
\end{minipage}
\end{center}

\noindent\textbf{ARC User Prompt -- Zero Shot}
\newline

\begin{center}
\noindent\begin{minipage}{0.95\linewidth}
\ttfamily\small
```\\Class Definitions:\\
no relation: The source and target arguments address different topics or aspects without directly supporting or contradicting each other.\\
\newline
attack: The source argument directly contradicts, refutes, or undermines the target argument.\\
\newline
support: The source argument provides evidence, reasoning, or additional points that strengthen or agree with the target argument.\\
\newline

{[}RELATION{]}: \{'no relation', 'attack', 'support'\}\\
{[}TOPIC{]}: <debate title>\\
{[}SOURCE{]}: <source argument>\\
{[}TARGET{]}: <target argument>\\'''
\end{minipage}
\end{center}

\noindent\textbf{ARC User Prompt -- Few Shot}

\begin{center}
\noindent\begin{minipage}{0.95\linewidth}
\ttfamily\small
```\\Class Definitions:\\
no relation: The source and target arguments address different topics or aspects without directly supporting or contradicting each other.\\
\newline
attack: The source argument directly contradicts, refutes, or undermines the target argument.\\
\newline
support: The source argument provides evidence, reasoning, or additional points that strengthen or agree with the target argument.\\

Example:\\
{[}SOURCE{]}: This regulation will protect consumers.\\
{[}TARGET{]}: We need stronger consumer protections.\\
{[}RELATION{]}: support\\

Example:\\
{[}SOURCE{]}: This is too costly for member states.\\
{[}TARGET{]}: The benefits far outweigh the costs.\\
{[}RELATION{]}: attack\\

{[}RELATION{]}: \{'no relation', 'attack', 'support'\}\\
{[}TOPIC{]}: <debate title>\\
{[}SOURCE{]}: <source argument>\\
{[}TARGET{]}: <target argument>\\
'''
\end{minipage}
\end{center}

\subsection*{Qualitative Example:}

Here we include a qualitative example to support the reading of Table \ref{tab:interpretation}, and demonstrate the potential of our proposed methods to facilitate interpretable evaluation. 
We focus on policy provision 15 ($x^p_{15}$ from Table \ref{tab:interpretation}) as it shows the largest variation in acceptability scores across debate summaries. 
The debate concerns a European Parliament resolution on the internal market and globalisation, and provision 15 reads:
 
\begin{quote}
\textit{``The European Parliament calls on the Commission and the Member States to continue the Better Regulation programme at both EU and national level and to report to the Parliament on progress.''}
\end{quote}
 
\noindent In the original debate $\mathcal{Q}$, $x^p_{15}$ has a strength of
$\sigma(\mathcal{Q}, x^p_{15}) = 0.47$, reflecting slight net opposition. However, in the Haiku summary, $\sigma(\mathcal{Q}^*_{\text{Haiku}}, x^p_{15}) = 0.89$, reflecting consensus approval, while the Sonnet summary reported $\sigma(\mathcal{Q}^*_{\text{Sonnet}}, x^p_{15}) = 0.54$, indicating a more balanced debate. To validate and contextualise these scores, we can explore the attacking and supporting arguments from each QBAF in detail:
 
\paragraph{Haiku Summary} ($\sigma(\mathcal{Q}^*_{\text{Haiku}}, x^p_{15}) = 0.89$)
 
\noindent \textbf{Support} ($(x^s_i, x^p_{15}) \in \mathcal{S}^*_{\text{Haiku}}$)
    \begin{itemize}
        \item \textit{``She argued that isolation is not an option and that the EU must
        ensure economic growth, employment, and competitiveness while supporting
        both large multinational companies and small and medium enterprises.''}
        \item \textit{``He stressed the importance of creating an attractive internal
        market, implementing the Lisbon Strategy, and developing more
        international regulatory convergence.''}
        \item \textit{``The consensus seemed to be that Europe must invest in education,
        research, innovation, and create flexible yet regulated market conditions
        to remain competitive on the global stage.''}
    \end{itemize}
\noindent \textbf{Attack} ($(x^s_i, x^p_{15}) \in \mathcal{A}^*_{\text{Haiku}}$)
    \begin{itemize}
        \item \textit{``Kyriacos Triantaphyllides criticised the report for potentially
        exacerbating social inequalities by supporting large market players.''}
    \end{itemize}
 
\paragraph{Sonnet Summary} ($\sigma(\mathcal{Q}^*_{\text{Sonnet}}, x^p_{15}) = 0.54$)
 
\noindent \textbf{Support} ($(x^s_i, x^p_{15}) \in \mathcal{S}^*_{\text{Sonnet}}$)
    \begin{itemize}
        \item \textit{``The report calls for better regulation in some over-regulated
        sectors and the creation of legal frameworks in areas lacking them,
        particularly regarding knowledge creation, intellectual property
        protection, and innovation.''}
        \item \textit{``He outlined key priorities including effective implementation of
        the Lisbon Strategy, completion of the internal market for services,
        telecommunications, energy and financial services, ensuring simple and
        effective regulation, and improving cross-border administrative
        cooperation.''}
        \item \textit{``Zuzana Roithov\'{a} characterised the report as a response to
        increasing problems facing the Union's internal market within the global
        economy.''}
        \item \textit{``She identified a major crossroads requiring Europe to review its
        social market economy regulations to increase flexibility while
        strengthening European influence on global economic rules.''}
        \item \textit{``She called for reviewing European regulation to support rather than
        hinder competitiveness, strengthening foreign policy, and asserting a
        values-based trade strategy.''}
    \end{itemize}
\noindent \textbf{Attack} ($(x^s_i, x^p_{15}) \in \mathcal{A}^*_{\text{Sonnet}}$)
    \begin{itemize}
        \item \textit{``Budreikait\.{e} criticised the hesitation of European Union states
        in making long-term economic decisions, particularly regarding energy and
        industrial policy.''}
        \item \textit{``Kyriacos Triantaphyllides argued the report is based on an
        erroneous perception promoted by the European Commission that contributed
        to the Constitutional Treaty's rejection.''}
        \item \textit{``He criticised the report for calling for increased internal market
        flexibility and supporting large European companies becoming global
        players, interpreting this as promoting greater employment flexibility
        and abolishing social state intervention.''}
        \item \textit{``Triantaphyllides cited statistics demonstrating extreme global
        inequality, noting that progress slowed during the period when markets
        globalised further, benefiting multinationals at the expense of small
        and medium enterprises.''}
        \item \textit{``Godfrey Bloom dismissed the idea of protecting failing national
        economies from global realities, arguing that state aid represents
        taxation and that money spent by politicians is nearly always wasted.''}
        \item \textit{``Andreas M\"{o}lzer argued that while globalisation cannot be
        stopped, the EU must create framework conditions to minimise victims
        within member states, attributing widespread fear to eastward enlargement
        and rapid globalisation.''}
    \end{itemize}
 
\noindent The Haiku summary retains three support arguments while reducing the entire opposition to a single attack, producing severe strength distortion. The Sonnet summary captures a substantially richer set of attack arguments (including the argument from $\mathcal{A}_{\text{Haiku}^*}$). This supports the reading of Table~4 and demonstrates that balance-consistency and $\epsilon$-accuracy are sensitive to degrees of failure rather than detecting
binary pass/fail outcomes.

\ifthenelse{\boolean{arxivversion}}{ }{\fi}

\end{document}